\documentclass[letterpaper, 10 pt, conference]{ieeeconf}  % Comment this line out if you need a4paper

\IEEEoverridecommandlockouts                              % This command is only needed if 
\usepackage{graphics} % for pdf, bitmapped graphics files
\usepackage{epsfig} % for postscript graphics files
\usepackage{mathptmx} % assumes new font selection scheme installed
\usepackage{times} % assumes new font selection scheme installed
\usepackage{amsmath} % assumes amsmath package installed
\usepackage{amssymb}  % assumes amsmath package installed

\usepackage{comment}
\usepackage{epstopdf}
\usepackage{subfigure}
\usepackage[table]{xcolor}
\usepackage{soul}
\usepackage[utf8]{inputenc}
\usepackage{caption}
\usepackage{flushend}

\definecolor{lightgray}{gray}{0.85}

\title{
\LARGE \bf 
Predicting Motion of Vulnerable Road Users \\
using High-Definition Maps and Efficient ConvNets
}

\author{
Fang-Chieh Chou, Tsung-Han Lin, Henggang Cui, Vladan Radosavljevic, \\
Thi Nguyen, Tzu-Kuo Huang, Matthew Niedoba, Jeff Schneider and Nemanja Djuric$^{1}$%
\thanks{$^{1}$Authors are with Uber Advanced Technologies Group, located at 3011 Smallman Street, Pittsburgh, PA, USA. Corresponding author e-mail: {\tt\small ndjuric@uber.com}}
}

\begin{document}

\maketitle
\thispagestyle{empty}
\pagestyle{empty}

%%%%%%%%% ABSTRACT
\begin{abstract}
Following detection and tracking of traffic actors, prediction of their future motion is the next critical component of a self-driving vehicle (SDV) technology, allowing the SDV to operate safely and efficiently in its environment. 
This is particularly important when it comes to vulnerable road users (VRUs), such as pedestrians and bicyclists. 
These actors need to be handled with special care due to an increased risk of injury, as well as the fact that their behavior is less predictable than that of motorized actors. 
To address this issue, in the current study we present a deep learning-based method for predicting VRU movement, where we rasterize high-definition maps and actor's surroundings into a bird's-eye view image used as an input to deep convolutional networks. 
In addition, we propose a fast architecture suitable for real-time inference, and perform an ablation study of various rasterization approaches to find the optimal choice for accurate prediction. 
The results strongly indicate benefits of using the proposed approach for motion prediction of VRUs, both in terms of accuracy and latency. 
%Following successful tests the system was deployed to a fleet of SDVs.

\end{abstract}

%%%%%%%%% BODY

\section{Introduction}
%Recent advances in high-performance hardware and software led to unprecedented breakthroughs in AI applications. %, with a number of highly publicized success stories that have captured attention and imagination of professionals and laymen alike. 
%Computers have surpassed human performance in centuries old games such as go \cite{silver2016mastering}, understand health conditions and suggest medical treatment \cite{topol2015patient}, and can even reason about complex relationships conveyed through images \cite{vinyals2017show}. This progress also prompted renewed enthusiasm and work on self-driving vehicles (SDVs), a nascent technology holding a potential to transform the way we live and work. While interest in SDVs goes as far back as the 1980s \cite{pomerleau1989alvinn}, only in the last decade government agencies and large industry players turned their focus towards the field, leading to a new era of research that caused leaps in real-world performance of SDVs \cite{urmson2008self}.

Predicting movement of traffic actors is a critical part of the autonomous technology. 
Once a self-driving vehicle (SDV) successfully detects and tracks a traffic actor in its vicinity, it needs to understand how will they move in the near future in order for both actors and SDV to be safe during operations \cite{dp2018}. 
This holds particularly true for vulnerable road users (VRUs), defined as traffic actors with increased risk of injury, unprotected by an outside shield \cite{oecd1998}. 
Road planners and policy makers have recognized this problem many decades ago, and attempted to mitigate it through several means. 
This included legal frameworks, designing new road types (e.g., segregating VRUs from motorized actors), educating both drivers and VRUs (with particular focus on children and elderly that are at an even greater risk than others \cite{oecd1998,10.1371/journal.pmed.1000228}), to name a few. 
These approaches have however given limited results, and in the US proportion of VRU deaths within overall traffic fatalities has actually increased between 2008 and 2017 from 14\% to 19\% \cite{ncsa2017} despite these best efforts.
%proportion of deaths of all outside-of-vehicle actors increased from 20\% to 33\% between 1996 and 2017, 
%while number of VRU fatalities in urban areas increased by staggering 46\% since 2008 \cite{ncsa2017}. To reverse the negative trend observed on our streets we arguably need a new approach to this old problem, and development of VRU-focused self-driving technology has a potential to do exactly that.

In the current study we address a critical aspect of the SDV technology, focusing on predicting future motion for VRUs, namely pedestrians and bicyclists. 
The main contributions of the paper are summarized below:
\begin{itemize}
  \item We present a system for motion prediction of VRU traffic actors, building upon recently proposed context rasterization techniques \cite{dp2018};
  \item We propose a fast and efficient convolutional neural network (CNN) architecture, suitable for running real-time onboard an SDV operating in crowded urban scenes with a large number of VRU actors;
  \item We present a detailed study of various rasterization settings identifying the optimal settings for accurate prediction, and provide critical insights into which parts of the system contribute the most to the accuracy;
  \item Following completion of offline tests the system was successfully tested onboard SDVs.
\end{itemize}

\section{Related work}
%Efficient and accurate detection, tracking, and motion prediction of VRUs are one of the key factors and requirements for autonomous vehicles to be safely deployed in complex urban environments. With greatly improved detection and tracking of VRUs \cite{Ohnbar2016}, research on motion prediction of VRUs has been gaining a lot of traction recently. Although most of the studies on the VRU motion prediction consider pedestrians with only a very few focusing on bicyclists, recent work showed importance of predicting bicyclists at crossings \cite{Cara2015} and signalized intersections \cite{Strauss2015MappingCA}. In this section we provide an overview of motion prediction of pedestrians and bicyclists as they relate to the problem of autonomous driving, while more comprehensive review that includes research on motion prediction of vehicles can be found in \cite{dp2018}.

Efficient and accurate motion prediction of VRUs is one of the key requirements to safely deploy SDVs in complex urban environments \cite{Ohnbar2016}. 
In this section we provide a literature overview of motion prediction of pedestrians and bicyclists in the context of autonomous driving.

\textbf{Motion prediction.}  
A common approach for prediction of VRU movement in autonomous driving systems is to use motion model from a tracking component to predict their future states. 
Tracking modules of most existing autonomous systems use either the Brownian or the constant velocity motion models \cite{Spinello2008MultimodalPD}. 
These models do not take into account scene contexts, and therefore fail in long-term prediction tasks as VRU motion follows complex patterns constrained by static and dynamic obstacles along the path. 
Traditionally, hand-crafted features were used for motion prediction of VRUs with respect to their surroundings. 
The social force model for pedestrian motion prediction incorporated interactive forces that guide pedestrians towards their goals and enforce collision avoidance among pedestrians, as well as between pedestrians and static obstacles \cite{Helbing1995, Yamaguchi2011}. 
Similar approach was applied for bicyclist motion prediction \cite{Huang2017}. In \cite{Pool2017} authors introduced a motion model for bicyclist motion prediction that incorporates knowledge of the road topology. 
The authors were able to improve prediction accuracy by using specific motion models for a pre-specified set of canonical directions. 
In \cite{trautman2015robot} interacting Gaussian Processes (GPs) with multiple goals were applied to model human cooperation in dense crowds for robot navigation. 
Authors of \cite{Habibi2018} predicted pedestrian trajectories by incorporating semantic scene features into the GP model, such as relative distance to curbside and state of traffic lights. 
A significant number of studies are devoted to modeling pedestrian motion using maximum entropy Inverse Reinforcement Learning (IRL) \cite{Ziebart2008}. 
In a followup work \cite{Ziebart2009} the authors introduced an IRL model based on a set of manually designed feature functions that capture interaction and collision avoidance behavior of pedestrians. 
While the approaches are capable of predicting pedestrian and bicyclist motions in many scenarios, the need for manual design of features makes them hard to scale in complex driving environments \cite{Varshneya2017}.

\textbf{Motion prediction using deep learning.} 
Inspired by their success in various areas of computer vision and robotics, many deep learning-based approaches were recently proposed for the motion prediction task in order to model object-object and object-scene interactions which may not be straightforward to represent manually.  
Most of deep learning approaches are based on Long Short-Term Memory (LSTM) variant of recurrent neural networks (RNNs) \cite{Hochreiter1997}. 
Authors of \cite{Sun2018} used a sequence-to-sequence LSTM encoder-decoder architecture to predict pedestrian position and heading. 
Incorporating angular information in addition to temporal features led to a significant improvement in accuracy. 
With respect to modeling of dynamic context, \cite{Alahi2016} proposed an LSTM-based approach for pedestrian motion prediction employing a ``social pooling'' layer that uses spatial information of nearby actors to implicitly model interactions among them. 
Vemula et al. \cite{Vemula2018} proposed ``social attention'' method to predict motion by estimating relative importance of pedestrians through an attention layer. 
Recently, \cite{Gupta_2018_CVPR} proposed an LSTM-based Generative Adversarial Network (GAN) to generate and predict socially feasible motions. 
On the other hand, with respect to modeling of static context \cite{Pfeiffer2018} proposed an LSTM-based model that incorporates the map of static obstacles and position of surrounding pedestrians. 
Moreover, \cite{Sadeghian2018} presented SoPhie, an LSTM-based GAN system for predicting physically and socially acceptable pedestrian trajectories using an RGB image from the scene and the trajectory information of surrounding actors. 
Similarly, in recent work \cite{Varshneya2017,Manh2018} authors incorporated scene information as well as human movement trajectories. 
In addition to LSTM-based methods, \cite{Nikhil2018} proposed CNN-based approach where convolutional layers are utilized to handle temporal dependencies. 
\cite{radwan2018multimodal} used an interaction-aware temporal CNN to predict pedestrian trajectories. 
\cite{chandra2019traphic} proposed a hybrid LSTM-CNN model that encodes actor states with LSTM and uses CNN to extract actor interaction features, while taking the varying behaviors of different road actors into account. 
\cite{luo2018fast} use CNNs for joint detection, tracking, and prediction. 
However, these models do not consider scene context, which can provide a strong signal on how the actors would move. 
On the other hand, in \cite{casas2018intentnet} the authors included map info to also predict high-level intent of vehicles, unlike our model that focuses on VRU actors instead. 

\textbf{Efficient CNN architectures.} 
%To apply deep neural networks on SDVs, we need to train networks that are both accurate and fast during inference. 
%Capability to give correct motion predictions in limited time budget is critical for SDVs to react quickly to abnormal road conditions and to prevent harmful accidents. 
%Modern SDVs are typically equipped with powerful GPU, therefore fast, real-time inference on GPU is highly sought-after. 
Since the introduction of seminal AlexNet \cite{alexnet}, researchers made significant progress in improving CNNs to make them more accurate and efficient. 
The state-of-the-art architectures, such as VGG \cite{vgg} or ResNet \cite{resnet}, tend to have a large number of layers running expensive computations, making them unsuitable for real-time inference. 
Recent proposals such as MobileNet \cite{MobileNet} and ShuffleNet \cite{ShuffleNet} replaced regular convolutional operator with a more efficient depthwise separable or group convolutions, making them small and fast for mobile applications. 
MobileNet-v2 (MNv2) \cite{MobileNetV2} further improved the original MobileNet by combining depthwise convolution with residual connections and bottleneck layers proposed in ResNet. 
One problem of this work is its focus on reducing number of floating point operations per second (FLOPS) instead of optimizing for actual latency on devices. 
More recently, MnasNet \cite{MnasNet}  applied search algorithms \cite{ZophL16} to optimize MNv2 architecture for both accuracy and inference latency on mobile devices, and is able to improve both while maintaining similar FLOPS. 
ShuffleNet v2 \cite{ShuffleNetV2} proposed several guidelines for designing fast networks beyond counting FLOPS, and applied these guidelines to design architectures suitable for both GPUs and mobile CPUs. 
In this paper we build upon and extend the MNv2 model, improving its speed on GPUs without compromising prediction accuracy.

\section{Proposed approach}

\begin{figure}[!t]
	\centering
	\includegraphics[keepaspectratio=1,width=0.28\textwidth]{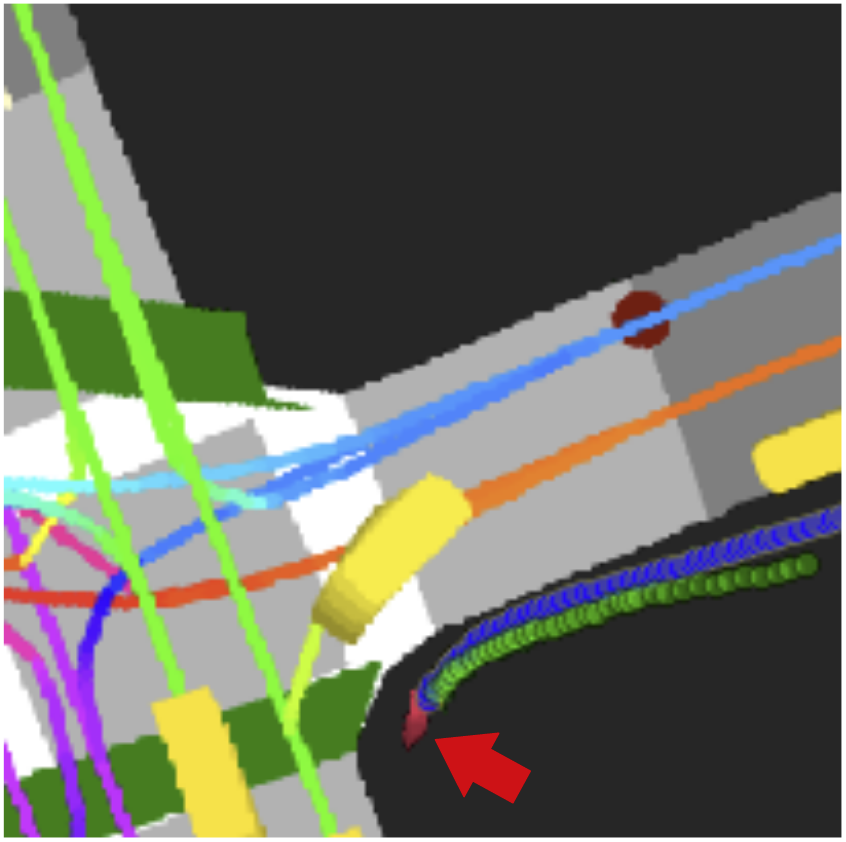}
    \caption{Example input raster for a pedestrian model with overlaid ground-truth (green) and output trajectory (blue); the target actor is colored in red and placed at the bottom-center of the raster image (indicated by the red arrow)}
	\label{fig:example_fig}
	\vspace{-.1in}
\end{figure}

%In this section we discuss our approach to trajectory prediction of VRU actors.
We build upon work described in \cite{dp2018}, which considered vehicle actors and used rasterized images of actor context as an input to CNNs to predict future trajectories.
In this study we extend the methodology to VRU actors (see Figure \ref{fig:example_fig} for an example of pedestrian motion prediction). 
Importantly, we improve the existing method's accuracy and inference speed by exploring two critical aspects. 
First, we experiment with different variations of the CNN architectures, and propose a novel architecture that significantly reduces inference latency without affecting accuracy; this is necessary to achieve real-time inference onboard an SDV in crowded urban environments comprising large number of VRUs. 
Second, we explore different rasterization configurations to find an optimal setup for highly accurate predictions of VRU actors.

\begin{figure*}[!t]
	\centering
	\includegraphics[keepaspectratio=1,width=0.65\textwidth]{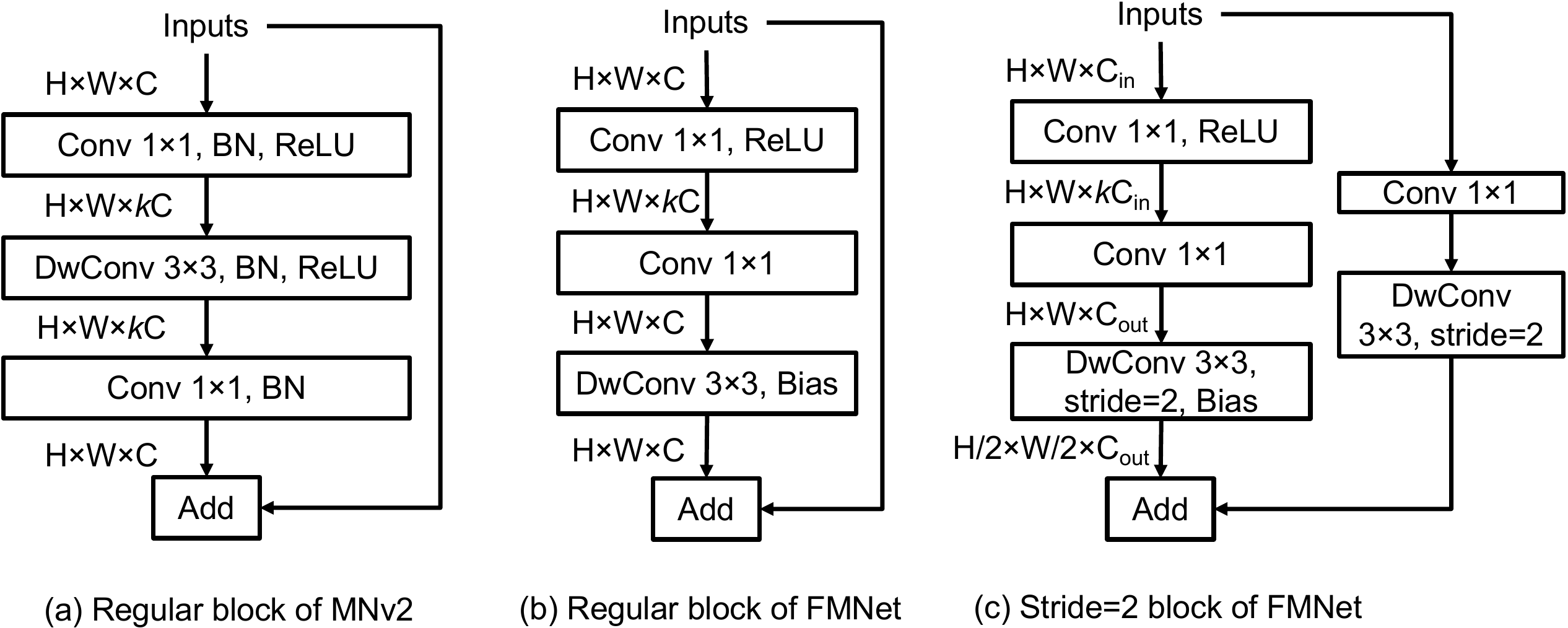}
	\caption{Building blocks of MobileNet-v2 \cite{MobileNetV2} and the proposed FastMobileNet (FMNet) architecture}
	\label{fig:fmnet_blocks}
	\vspace{-.1in}
\end{figure*}

Let us assume we have access to real-time data streams coming from sensors such as lidar, radar, or camera, installed aboard an SDV. 
In addition, assume these inputs are used by an existing detection and tracking system, outputting state estimates $\mathcal{S}$ for all surrounding actors (state comprises the bounding box, position, velocity, acceleration, heading, and heading change rate). 
Denote a set of discrete times at which tracker outputs state estimates as $\mathcal{T} = \{t_1, t_2, \dots, t_T\}$, where time gap between consecutive time steps is constant (e.g., the gap is equal to $0.1s$ for tracker running at a frequency of $10Hz$). 
Then, we denote state output of a tracker for the $i$-th actor at time $t_j$ as ${\bf s}_{ij}$, where $i = 1, \dots, N_j$ with $N_j$ being a number of unique actors tracked at $t_j$. 
%Note that in general the actor counts vary for different time steps as new actors appear within and existing ones disappear from the sensor range. 
Moreover, we assume access to a detailed, high-definition map $\mathcal{M}$ of the SDV's operating area, including road and crosswalk locations, lane directions, and other relevant map information.

Using the state estimates and high-definition map, for each actor of interest we rasterize an actor-specific bird's-eye view raster image encoding the actor's surrounding map and traffic actors, as illustrated in Figure \ref{fig:example_fig}. 
Then, given the $i$-th actor's raster image at time step $t_j$ and state estimate ${\bf s}_{ij}$, we use a CNN model to predict a sequence of its future states up to the prediction horizon of $H$ time steps $[{\bf s}_{i(j+1)}, \dots, {\bf s}_{i(j+H)}]$ (see model architecture in Figure \ref{fig:fusion_arch}a), trained to minimize the average displacement error (ADE) of the predicted trajectory points. 
Without loss of generality, in this work we simplify the task to infer the $i$-th actor's future $x$- and $y$-positions instead of full state estimates, while the remaining states can be derived from the current state ${\bf s}_{ij}$ and the future predicted position estimates. 
Both past and future positions at time $t_j$ are represented in the actor-centric coordinate system derived from actor's state at time $t_j$, where forward direction is $x$-axis, left-hand direction is $y$-axis, and actor's bounding box centroid represents the origin.

Following the approach in \cite{dp2018} we use an MNv2 model as the base CNN to compute future positions from the input raster image. 
Below we describe improvements to this architecture, followed by discussion of variations to the rasterization process that were considered in the study.

\begin{table*} [!ht]
\caption{Architecture of FastMobileNet (upsample factor for all FMNet blocks is set to $k=6$)}
\label{tab:fmnet_arch}
\centering
{
\begin{tabular}{cccc}
	{\bf Layer} & {\bf Output size} & {\bf Stride} & {\bf Repeats} \\
	\hline
        \rowcolor{lightgray}
	Raster image & $300 \times 300 \times 3$ & $-$ & $-$  \\
	Conv $3 \times 3$ & $150 \times 150 \times 24$ & 2 & 1 \\
        \rowcolor{lightgray}
	DwConv $3 \times 3$ & $75 \times 75 \times 24$ & 2 & 1 \\
	FMNet block 1 & $75 \times 75 \times 12$ & 1 & 2 \\
        \rowcolor{lightgray}
	FMNet block 2 & $38 \times 38 \times 16$ & 2 & 3 \\
	FMNet block 3 & $19 \times 19 \times 32$ & 2 & 4 \\	
        \rowcolor{lightgray}
	FMNet block 4 & $19 \times 19 \times 48$ & 1 & 3 \\	
	FMNet block 5 & $10 \times 10 \times 80$ & 2 & 3 \\	
        \rowcolor{lightgray}
	FMNet block 6 & $10 \times 10 \times 160$ & 1 & 1 \\
	Conv $1 \times 1$ & $10 \times 10 \times 640$ & 1 & 1 \\
        \rowcolor{lightgray}
	Global average pooling & $1 \times 1 \times 640$ & 1 & 1 \\	
	\hline
\end{tabular}
}
\end{table*} 

\subsection{Improved CNN architecture for fast inference}

\subsubsection{Base CNN}
In this section we propose several modifications to the MNv2 architecture that lead to significant speedup of GPU inference, making it feasible to perform real-time inference when our SDVs operate in urban scenes containing large number of VRUs. 
MNv2 is based on the inverted bottleneck block illustrated in Figure \ref{fig:fmnet_blocks}a. 
In each block, the input feature map is first upsampled to $k$ times more channels with $1 \times 1$ convolutions ($k$ is set to 6 in the original MNv2), followed by $3 \times 3$ depthwise convolution (DwConv) applied to the upsampled feature map. 
Then, the feature map is compressed back to the original channel size using $1 \times 1$ convolution, and summed with the initial input through residual connection. Non-linear activation function (e.g., ReLU) is applied only in the upsampled phase, as non-linearity in the bottlenecked phase (before the upsampling or after the compression) causes too much information loss and hurts model performance. 
BatchNorm (BN) is used in all three layers. 
While the majority  of the FLOPS are in the $1 \times 1$ convolutions (amounting to $87 \%$ of the total), the other operations still incur non-negligible cost. 
As discussed in \cite{ShuffleNetV2}, FLOPS itself is not an accurate metric of latency, and another important factor is the number of memory access operations (MAC). 
Operations such as DwConv, BatchNorm, ReLU, and BiasAdd, while having small FLOPS, typically incur heavy MAC. 
This especially holds true for MNv2, where operations in the upsampled phase have $k$ times more MAC than the same operations in the bottlenecked phase. 

\begin{figure*}[!t]
	\centering
	\includegraphics[keepaspectratio=1,width=0.75\textwidth,trim={0 2.0in 0 0},clip]{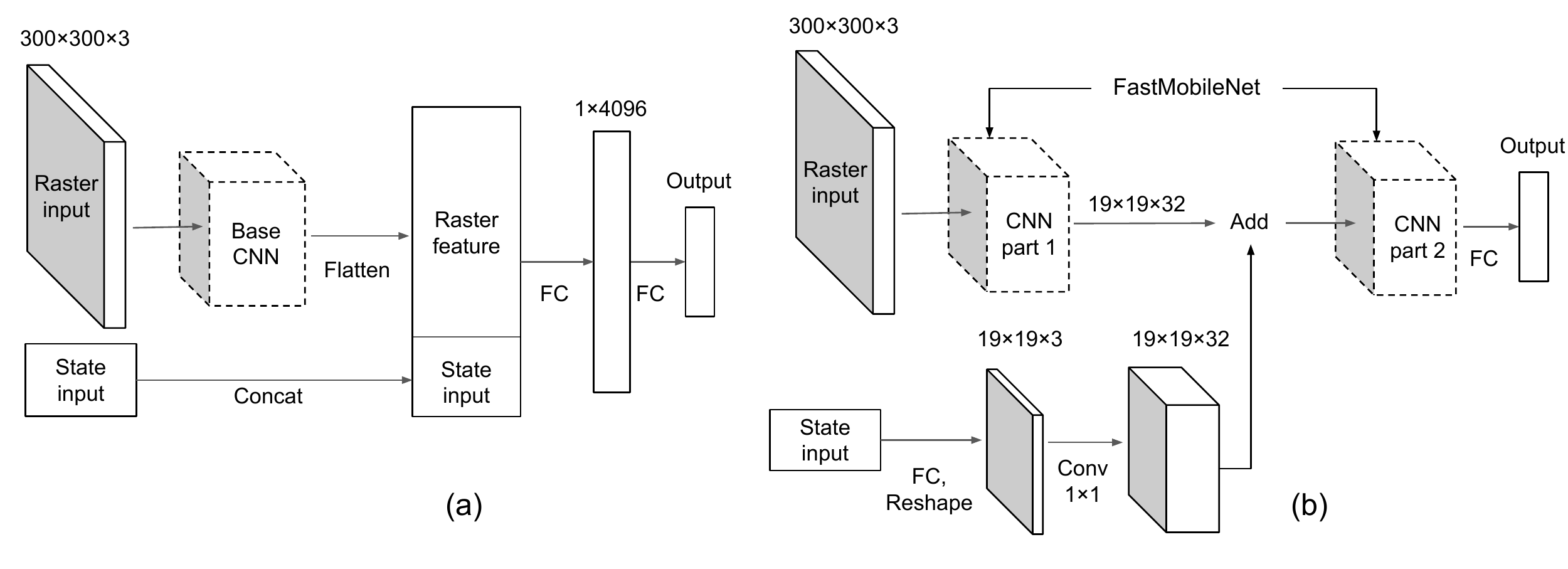}
	\caption{Feature fusion through (a) concatenation; and (b) spatial fusion}
	\vspace{-.05in}
	\label{fig:fusion_arch}
\end{figure*}

Compared to MNv2, in the proposed novel CNN architecture called FastMobileNet (FMNet) we move most of the operations originally in the upsampled phase into the bottlenecked phase, reducing their FLOPS and MAC by a factor of $k$, as illustrated in Figure \ref{fig:fmnet_blocks}b. 
Note that we show the architecture for an equal number of input and output channels, if they are not the same an additional $1 \times 1$ convolution is applied after the residual connection.
The only remaining operation in the upsampled phase is a ReLU. 
Similarly to MNv2, no ReLU is applied in the bottlenecked phase since applying non-linearity there causes significant information loss. 
The layers are linear in the bottlenecked phase, and we only apply one BiasAdd at the end of the block as applying multiple BiasAdd in consecutive linear layers does not increase model expressiveness. 
We do not use BatchNorm in FMNet as we found that the model converges well without it, and excessive BatchNorms cost additional computation time. 
In addition, we need to allow different strides in order to reduce height and width of the feature map during feature extraction. 
The FMNet block supporting this operation is similar to the regular block (see Figure \ref{fig:fmnet_blocks}c), except the original input is downsampled to the correct output size for residual connection. 
The base model (further discussed and extended in the following section) is illustrated in Figure \ref{fig:fusion_arch}a, and the FMNet architecture corresponding to the CNN part is shown in Table \ref{tab:fmnet_arch}, where the layer sizes and block repeats of the model are based on MNv2-0.5 \cite{MobileNetV2} (i.e., MNv2 with halved channel sizes in all layers). 

\subsubsection{Fusion of auxiliary features}
Previous work \cite{dp2018} showed that combining the raster input with other state features of actors (e.g., current and past velocity, acceleration, heading change rate) significantly improves model accuracy. 
Thus, it is beneficial to design a network that fuses the raster image input (as a 3D-tensor of size $height \times width \times channels$) and other auxiliary features (as a 1D-vector) that include the actor states and/or other hand-engineered features. 
A straightforward way to achieve this, as done in \cite{dp2018}, is to concatenate the flattened CNN output from the raster image with the 1D auxiliary features, then apply additional fully-connected layers to allow non-linear feature interactions, as shown in Fig.~\ref{fig:fusion_arch}a. 
In this section we propose an alternative, more efficient way to fuse the raster CNN and the auxiliary features. 
We convert the 1D auxiliary features into a 3D feature map by a sequence of a fully-connected layer, reshaping, and $1 \times 1$ convolution, and fuse it into an intermediate CNN feature map by element-wise addition, as illustrated in Fig.~\ref{fig:fusion_arch}b. 
In this way, we reuse existing downstream CNN computations to achieve non-linear interactions between raster features and the auxiliary features. 
This removes a need for the additional fully-connected layer that is used in the fusion through concatenation, thus saving valuable computation time. 
Furthermore, in this way we allow the feature pixels at different spatial locations of the CNN feature map to interact differently with auxiliary features. 
We perform feature fusion at the output of FMNet block 3 (see Table~\ref{tab:fmnet_arch}). 
As discussed in the evaluation section, we found that this spatial feature fusion method leads to improved model accuracy and latency.

\subsection{Exploring various rasterization settings}
\label{sect:rasterization}
To describe rasterization, let us first introduce a concept of a {\it vector layer}, formed by a collection of polygons and lines that belong to a common type. 
For example, in the case of map elements we may have vector layer of roads, of crosswalks, and so on. 
To rasterize vector layer into an RGB space, each vector layer is manually assigned a color from a set of distinct RGB colors that make a difference among layers more prominent. 
Once the colors are defined, vector layers are rasterized one by one on top of each other in the order from layers that represent larger areas, such as road polygons, towards layers that represent finer structures, such as lanes or actor bounding boxes.
To represent context around the $i$-th actor tracked at time step $t_j$ we create a rasterized image $I_{ij}$ of size $n \times n$ such that the actor is positioned at pixel $(w, h)$ within $I_{ij}$, where $w$ represents width and $h$ represents height measured from the bottom-left corner of the image, with actor heading always pointing up. 
We color the actor of interest differently so that it is distinguishable from other actors. 
See Figure \ref{fig:example_fig} for an example raster with pedestrian actor of interest, while more detailed explanation of rasterization can be found in \cite{dp2018}.

\begin{figure*}[!ht]
	\centering
	\includegraphics[keepaspectratio=1,width=0.18\textwidth]{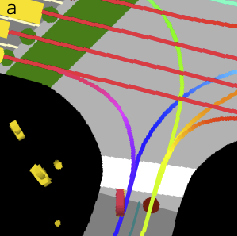}
	\includegraphics[keepaspectratio=1,width=0.18\textwidth]{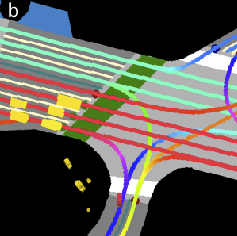}
	\includegraphics[keepaspectratio=1,width=0.18\textwidth]{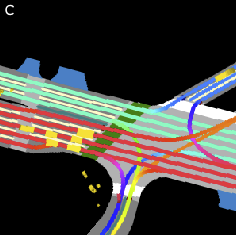}
	\caption{Raster images for bicyclist actor (colored red) using resolutions of $0.1m$, $0.2m$, and $0.3m$, respectively}
	\label{fig:raster_resolution}
\end{figure*}

\begin{figure*}[!ht]
	\centering
	\includegraphics[keepaspectratio=1,width=0.18\textwidth]{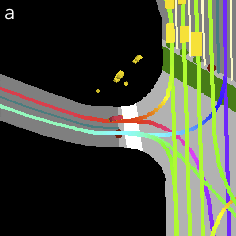}
	\includegraphics[keepaspectratio=1,width=0.18\textwidth]{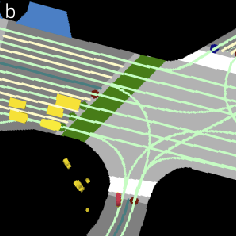}
	\includegraphics[keepaspectratio=1,width=0.18\textwidth]{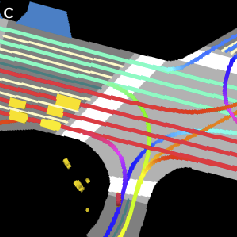}
	\includegraphics[keepaspectratio=1,width=0.18\textwidth]{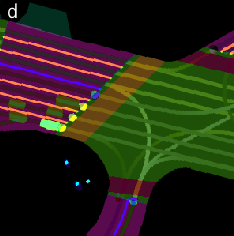}
	\caption{Different rasterization settings with $0.2m$ resolution for a bicyclist example: (a) no raster rotation, (b) no lane heading encoding, (c) no traffic light encoding, (d) learned colors}
	\label{fig:raster_no_X}
	\vspace{-.1in}
\end{figure*}

In this study we evaluated several different choices of rasterization for the motion prediction of VRU actors, and their impact on the model performance. 
For all approaches we maintain a constant RGB raster dimension of $300 \times 300$ pixels (i.e., we set $n=300$), and discuss the specifics of various choices below.

\textbf{Raster pixel resolution.} 
The resolution governs the extent of surrounding context seen by the model. 
At $0.1m$ resolution, the model sees $25m$ in front and $5m$ behind the actor (assuming the image size discussed above). 
Larger resolution allows for larger context around the actor to be captured, however the raster loses finer details which may be critical for accuracy. 
To evaluate its impact we experimented with resolutions of $0.1m$, $0.2m$ and $0.3m$, as shown in Figure \ref{fig:raster_resolution}.

\textbf{Raster frame rotation.} 
During rasterization we can rotate each raster separately per actor \cite{dp2018}, such that the actor heading points up and the target actor is placed at $w=150, h=50$ (as seen in Figure~\ref{fig:raster_resolution}). 
In this way actor heading is encoded directly into the input, and the raster captures more context in front of the actor. 
We tested an alternative scheme where the raster frame is not rotated such that upward direction indicates north instead, and the actor is placed in the center (setting $w=h=150$), as seen in Figure~\ref{fig:raster_no_X}a.

\textbf{Lane direction.} 
As proposed in earlier work \cite{dp2018}, the direction of each lane segment can be encoded as a hue value in HSV color space with saturation and value set to a maximum, followed by the conversion of HSV to RGB color space. 
Alternatively we can encode all lanes with a constant color, such that the raster does not contain information on lane direction. 
This is represented in Figure \ref{fig:raster_no_X}b where raster does not encode lane direction, as opposed to Figure \ref{fig:raster_resolution}b where lane color indicates its heading.

\textbf{Traffic lights.} 
We use an existing in-house traffic light classification algorithm to extract current traffic light states from sensor inputs. 
To encode this info in the input raster, we plot traffic light states as a colored circle at location where lane meets a traffic-light controlled intersection. 
Furthermore, we identify inactive crosswalks and paint them green, signaling that vehicles may pass through a crosswalk (compare Figure \ref{fig:raster_no_X}c where raster image does not encode traffic light info to Figure \ref{fig:raster_resolution}b where it does).

\textbf{Learning raster colors.}
When generating raster images the colors for each raster layer type can be chosen manually, as proposed in \cite{dp2018}.
An alternative approach is to have the DNN learn the colors by itself, optimizing raster image for the prediction task.
In this study we provide all raster layers (e.g., road and crosswalk polygons, tracked objects) to the network as separate binary-valued channels, and add a $1\times1$ convolution layer with 3 output channels and linear activation to generate the RGB raster image (see Figure \ref{fig:raster_no_X}d for an example of a learned raster).
The resulting RGB image is then passed to the rest of the network as before.

\textbf{Model pre-training.} 
Lastly, we evaluated one modification that is not related to rasterization choices. 
As the majority of actors observed on roads are vehicles, our training data has a much larger number of such traffic actors. 
To make use of this data, we can initialize our VRU models with a pre-trained vehicle model trained using more examples. 
The models can then be fine-tuned using corresponding VRU training examples until convergence.

\section{Experiments}
\label{sect:exp}

\begin{table*} [!t]
\caption{Comparison of various CNN architectures (all models except the last one use the concatenation feature fusion)}
\label{tab:architecture}
\centering
{
\begin{tabular}{ccccccc}
    {\bf Architecture} & {\bf ADE [m]} & {\bf Latency [ms]} & {\bf FLOPS} & {\bf Num. parameters} & {\bf MAC} & {\bf Num. ops} \\
    \hline
    \rowcolor{lightgray}
    AlexNet & 1.36 & 15.8 & 2.63G & 70.3M & 364 MB & {\bf 131} \\
    ResNet18 & 1.29 & 36.2 & 6.26G & 11.7M & 163 MB & 641 \\
    \rowcolor{lightgray}
    MNv2-0.5 & 1.27 & 21.3 & 308M & 598K & 146 MB & 1542 \\
    MnasNet-0.5 & 1.28 & 18.3 & 323M & 844K & 113 MB & 1490 \\
    \rowcolor{lightgray}
    FMNet & 1.28 & 12.1 & 340M & 565K & 55 MB & 336 \\
    FMNet with spatial fusion & {\bf 1.24} & {\bf 10.4} & {\bf 285M} & {\bf 558K} & {\bf 47 MB} & 370 \\
    \hline
\end{tabular}
}
\end{table*}

\begin{table*} [!t]
\caption{Comparison of prediction displacement errors (in meters) for different experimental settings}
%\vspace{-.05in}
\label{tab:rasterization}
\centering
{
  \begin{tabular}{cccccccccccc}
     & & \multicolumn{3}{c}{\bf Bicyclists} & \multicolumn{3}{c}{\bf Pedestrians} \\
    {\bf Approach} & {\bf Resolution} & {\bf Average} & {\bf @1s} & {\bf @5s} & {\bf Average} & {\bf @1s} & {\bf @5s} \\
    \hline
    \rowcolor{lightgray}
    UKF & $-$ & 2.89 & 0.80 & 6.60 & 0.67 & 0.22 & 1.22 \\
    Social-LSTM & $-$ & 3.79 & 1.85 & 6.61 & 0.53 & 0.29 & 0.95 \\
    \hline
    \rowcolor{lightgray}
    RasterNet & $0.1m$ & 1.07 & 0.43 & 2.73 & {\bf 0.51} & {\bf 0.17} & {\bf 0.90} \\
    RasterNet & $0.2m$ & 1.07 & 0.44 & 2.72 & 0.52 & 0.18 & 0.93 \\
    \rowcolor{lightgray}
    RasterNet & $0.3m$ & 1.09 & 0.45 & 2.80 & 0.53 & 0.18 & 0.95 \\
    RasterNet w/o rotation & $0.2m$ & 1.29 & 0.49 & 3.30 & 0.58 & 0.20 & 1.02 \\
    \rowcolor{lightgray}
    RasterNet w/o  traffic lights & $0.2m$ & 1.11 & 0.44 & 2.86 & 0.55 & 0.20 & 0.96 \\
    RasterNet w/o  lane headings & $0.2m$ & 1.07 & 0.43 & 2.72 & 0.52 & 0.18 & 0.93 \\
    \rowcolor{lightgray}
    RasterNet with learned colors & $0.2m$ & {\bf 1.05} & {\bf 0.42} & {\bf 2.70} & 0.53 & 0.18 & 0.93 \\
    \hline
    RasterNet vehicle model & $0.2m$ & 3.11 & 0.89 & 8.47 & 1.96 & 0.40 & 3.82  \\
    \rowcolor{lightgray}
    RasterNet vehicle fine-tuned & $0.2m$ & {\bf 1.05} & {\bf 0.42} & {\bf 2.70} & 0.59 & 0.20 & 1.05 \\
    \hline
\end{tabular}
}
\end{table*}

We collected 240 hours of data by manually driving SDV in various traffic conditions (e.g., varying times of day, days of the week). 
The data contains significantly different number of examples for various actor types, namely 7.8 million vehicles, 2.4 million pedestrians, and 520 thousand bicyclists. 
Traffic actors were tracked using Unscented Kalman filter (UKF) \cite{wan2000unscented}, taking raw sensor data from the camera, lidar, and radar, and outputting state estimates for each object at $10Hz$. 
%The filter is a default tracker on our fleet, trained on a large amount of labeled data, and tested on millions on miles (unfortunately, no other details can be given due to confidentiality concerns). 
We considered prediction horizon of $6s$ (i.e., $H = 60$) for VRU actors. 
For the default rasterization scheme (used in the architecture experiments and as a base setting in the rasterization ablation study), we rotated raster to actor frame with resolution of $0.2m$, including in the raster image both lane heading and traffic light layers (illustrated in Figure \ref{fig:raster_resolution}b).

We implemented models in TensorFlow \cite{tensorflow2015-whitepaper} and trained on 16 Nvidia GTX 1080Ti GPU cards. 
We used open-source distributed framework Horovod \cite{sergeev2018horovod} for training, completing in around 24 hours. 
We used a per-GPU batch size of $64$ and Adam optimizer \cite{kingma2014adam}, setting initial learning rate to $10^{-4}$ further decreased by a factor of $0.9$ every $20{,}000$ iterations.

\subsection{Comparison of CNN architectures}
In the first set of experiments we compared a number of CNN architectures, summarizing results in Table \ref{tab:architecture}. 
To ensure fair comparison and avoid potential issues with small data sets, we trained all models on vehicle actors where we set the prediction horizon to $6s$. 
Average displacement error and latency are reported in the table. 
We skip the feature fusion layers when computing the number of parameters, as the concatenation feature fusion adds a large amount of parameters which complicates the comparison (an $1024 \times 4096$ fully-connected layer for feature fusion adds 4M extra parameters). 
MAC is approximated by the sum of tensor sizes of all graph nodes. Column ``{\it Num. ops}" refers to the total number of operations in the TensorFlow graph of each model. 
The inference latency is measured at a batch of 32 actors on a GTX 1080Ti GPU. 
As our prediction algorithm performs inference for each actor in the scene, having such a large batch size is not uncommon when SDV is driving on crowded streets. 
Note that model latency is implementation-specific, as fusing graph operations manually with TensorFlow custom ops or automatically using Nvidia TensorRT might affect latency. 
For simplicity and to facilitate fair comparison, we implemented all models using TensorFlow built-in operations without additional optimization.

First, we compared the prediction accuracy and inference latency on several base CNN architectures. 
We found that the proposed FMNet gives similar prediction accuracy as other modern architectures (such as ResNet, MNv2, and MnasNet), while being much faster during inference. 
In terms of the number of FLOPS and parameters, FMNet is similar to MnasNet-0.5 and MNv2-0.5 which it is based on, while AlexNet and ResNet18 are much more complex. 
Fast inference of FMNet can be explained by low MAC and operation counts, which we specifically optimized for during the model design phase. 
It is interesting to note that AlexNet is the second fastest CNN while having the second largest FLOPS. 
This is due to it having the smallest number of layers, as evidenced by its lowest operation counts, although its accuracy is not on par with the competing networks.

Secondly, we found that FMNet with spatial feature fusion further improves the accuracy and inference time when compared to the model with feature fusion through concatenation. 
As discussed in the previous section, the spatial fusion allows interactions between raster and state features with awareness of spatial locations, and removes an expensive fully-connected layer used in the original architecture. 
This resulted in lower FLOPS, as well as lower number of parameters and MAC.
Following these results we use the best performing FMNet with spatial fusion as the model architecture in the following study on rasterization choices.

\begin{figure*}[!t]
	\centering
	\includegraphics[keepaspectratio=1,width=0.29\textwidth]{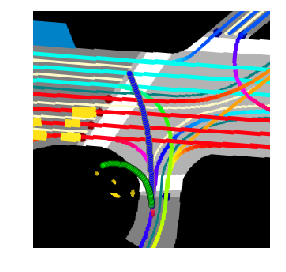}
	\includegraphics[keepaspectratio=1,width=0.29\textwidth]{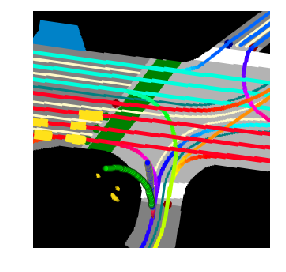}
	\caption{Bicyclist model before and after traffic light turns red; ground-truth (green) and predicted (blue) trajectories overlaid}
	\label{fig:inference_tl}
	\vspace{-.1in}
\end{figure*}

\subsection{Comparison of prediction models and inputs}
We compared the proposed method to the state-of-the-art baselines, and conducted an ablation study analyzing different rasterization setups to identify an optimal setting for accurate VRU predictions. 
Empirical results in terms of average displacement error, as well as short- and long-term displacement errors are given in Table~\ref{tab:rasterization}. 

As the baselines we considered a simple rollout using the UKF, as well as Social-LSTM \cite{Alahi2016} trained on our data. 
We also tried Social-GAN \cite{Gupta_2018_CVPR}, but it did not converge on our data set and is thus not shown here. 
We can see that our CNN with optimized architecture (referred to as {\it RasterNet}) outperformed UKF and Social-LSTM, due to its encoding of surrounding context and actors in the input raster. 
Social-LSTM gives competitive prediction result for pedestrians since it handles actor interactions in the network, but fails to give accurate prediction on bicyclists. This is possibly due to the lack of surrounding context information in its input (e.g., the lane graph and traffic signal states), which is critical for accurate prediction of bicyclist motion.

For the rasterization ablation study, we first analyzed accuracy of the base $0.2m$-resolution, as compared to other resolution choices. 
Resolution of $0.1m$ has smaller coverage of the surrounding area, and is expected to benefit slow-moving objects such as pedestrians. 
On the other hand, resolution of $0.3m$ may benefit faster-moving objects requiring larger coverage, while also resulting in a loss of finer details useful for slower-moving actors. 
As can be seen, $0.1m$-resolution indeed resulted in lower error for pedestrians, while setting $0.3m$ gave the worst performance. 
We observed that resolution of $0.1m$ showed no significant difference for bicyclists, while $0.3m$ resulted in slightly higher error. 
This may be attributed to the fact that bicyclists are not fast enough to benefit from wider context.

Next, we evaluated the impact of not rotating raster such that actor heading points up, as discussed in Section \ref{sect:rasterization}, which resulted in a significant drop of accuracy for both actor types. 
This can be explained by the fact that when raster is not rotated there is a large number of input data variations that network needs to observe in order to learn how actors move, and the input data could be augmented by randomly rotating each example. 
In other words, for such setup actors may initially move in any direction, which is not the case for the rotated raster where actors initially always move upward, resulting in a simplified prediction problem.

We further investigated the affect of encoding traffic light info. 
The traffic light is important for predicting longitudinal movement, as it can provide info about whether an actor may or may not pass through an intersection. 
We observed error increase without traffic light rasterization for both actor types, matching this intuition. 
Figure \ref{fig:inference_tl} gives an illustrative example of how additional traffic information improves prediction, where the bicyclist's predicted trajectory changed from crossing to non-crossing due to the different traffic light state. 
In this case the ground-truth trajectory is indeed non-crossing, where the actor moved into the sidewalk to wait for the vehicle traffic to pass. 
Furthermore, we removed lane heading info provided in rasters, encoded by using different colors to indicate different lane directions. 
Without lane heading the bicyclist model degraded slightly in performance, while the pedestrian model was unaffected. 
This matches the intuition, as bicyclists may behave as vehicles at times and follow lane direction, while pedestrians do not use lane heading when in traffic.
Finally, we tried learning raster colors instead of setting them manually. 
The results indicate that learned colors slightly improved accuracy for bicyclists, whereas the pedestrian model slightly degraded compared to the baseline. 
This suggests that our manual rasterization setup captured sufficient signal when it comes to motion prediction of pedestrians, while for bicyclists it could be suboptimal.

Lastly, due to significantly larger amount of vehicle data as compared to VRUs, instead of training from scratch we fine-tuned VRU models using preloaded weights from a pretrained vehicle model. 
As observed in the second-to-last row, directly applying the vehicle model to bicyclists and pedestrians without fine-tuning results in poor performance across the board, even worse than UKF, due to a different nature of these actor types as compared to vehicles. 
On the other hand, with additional fine-tuning using training data of corresponding actor types the bicyclist performance improved over the baseline, indicating that bicyclists may exhibit somewhat similar behavior to vehicles. 
We can also see that the pedestrian model regressed, explained by the fact that pedestrian motion is very different from vehicle motion, thus making vehicle pretraining ineffective.

\section{Conclusion}
%We presented an efficient and effective solution to 
Motion prediction of VRU actors is a critical problem in autonomous driving, as such actors have higher risk of injury and are less predictable. % since they may change behavior faster than vehicles. 
We applied recently proposed rasterization technique to generate raster images of actors' surroundings, used as inputs to CNNs trained to predict actor trajectory. Importantly, we proposed a fast architecture suitable for real-time SDV operations in crowded urban scenes, and conducted a detailed ablation study of rasterization settings to identify the optimal choice for accurate VRU predictions. 
The results indicate benefits of the proposed approaches, and provide useful insights into the task of motion prediction. 
Moreover, following extensive offline tests the model was successfully tested onboard SDVs.

{\small
\bibliographystyle{IEEEtran}
\bibliography{references}

\begin{thebibliography}{10}
\providecommand{\url}[1]{#1}
\csname url@rmstyle\endcsname
\providecommand{\newblock}{\relax}
\providecommand{\bibinfo}[2]{#2}
\providecommand\BIBentrySTDinterwordspacing{\spaceskip=0pt\relax}
\providecommand\BIBentryALTinterwordstretchfactor{4}
\providecommand\BIBentryALTinterwordspacing{\spaceskip=\fontdimen2\font plus
\BIBentryALTinterwordstretchfactor\fontdimen3\font minus
  \fontdimen4\font\relax}
\providecommand\BIBforeignlanguage[2]{{%
\expandafter\ifx\csname l@#1\endcsname\relax
\typeout{** WARNING: IEEEtran.bst: No hyphenation pattern has been}%
\typeout{** loaded for the language `#1'. Using the pattern for}%
\typeout{** the default language instead.}%
\else
\language=\csname l@#1\endcsname
\fi
#2}}

\bibitem{dp2018}
N.~Djuric, V.~Radosavljevic, H.~Cui, T.~Nguyen, F.-C. Chou, T.-H. Lin,
  N.~Singh, and J.~Schneider, ``Uncertainty-aware short-term motion prediction
  of traffic actors for autonomous driving,'' \emph{IEEE Winter Conference on
  Applications of Computer Vision (WACV)}, 2020.

\bibitem{oecd1998}
OECD, ``Safety of vulnerable road users,'' Organisation for Economic
  Co-operation and Development, Tech. Rep. 68074, Augist 1998.

\bibitem{10.1371/journal.pmed.1000228}
\BIBentryALTinterwordspacing
A.~Constant and E.~Lagarde, ``Protecting vulnerable road users from injury,''
  \emph{PLOS Medicine}, vol.~7, no.~3, pp. 1--4, 03 2010. [Online]. Available:
  \url{https://doi.org/10.1371/journal.pmed.1000228}
\BIBentrySTDinterwordspacing

\bibitem{ncsa2017}
NCSA, ``2017 fatal motor vehicle crashes: Overview,'' National Center for
  Statistics and Analysis, Tech. Rep. DOT HS 812 603, October 2018.

\bibitem{Ohnbar2016}
E.~Ohn-Bar and M.~M. Trivedi, ``Looking at humans in the age of self-driving
  and highly automated vehicles,'' \emph{IEEE Transactions on Intelligent
  Vehicles}, vol.~1, no.~1, pp. 90--104, March 2016.

\bibitem{Spinello2008MultimodalPD}
L.~Spinello, R.~Triebel, and R.~Siegwart, ``Multimodal people detection and
  tracking in crowded scenes,'' in \emph{AAAI}, 2008.

\bibitem{Helbing1995}
\BIBentryALTinterwordspacing
D.~Helbing and P.~Moln\'ar, ``Social force model for pedestrian dynamics,''
  \emph{Phys. Rev. E}, vol.~51, pp. 4282--4286, May 1995. [Online]. Available:
  \url{https://link.aps.org/doi/10.1103/PhysRevE.51.4282}
\BIBentrySTDinterwordspacing

\bibitem{Yamaguchi2011}
K.~Yamaguchi, A.~C. Berg, L.~E. Ortiz, and T.~L. Berg, ``Who are you with and
  where are you going?'' in \emph{CVPR 2011}, June 2011, pp. 1345--1352.

\bibitem{Huang2017}
L.~Huang, J.~Wu, F.~You, Z.~Lv, and H.~Song, ``Cyclist social force model at
  unsignalized intersections with heterogeneous traffic,'' \emph{IEEE
  Transactions on Industrial Informatics}, vol.~13, no.~2, pp. 782--792, April
  2017.

\bibitem{Pool2017}
E.~A.~I. Pool, J.~F.~P. Kooij, and D.~M. Gavrila, ``Using road topology to
  improve cyclist path prediction,'' in \emph{2017 IEEE Intelligent Vehicles
  Symposium (IV)}, June 2017, pp. 289--296.

\bibitem{trautman2015robot}
P.~Trautman, J.~Ma, R.~M. Murray, and A.~Krause, ``Robot navigation in dense
  human crowds: Statistical models and experimental studies of human--robot
  cooperation,'' \emph{The International Journal of Robotics Research},
  vol.~34, no.~3, pp. 335--356, 2015.

\bibitem{Habibi2018}
G.~Habibi, N.~Jaipuria, and J.~P. How, ``Context-aware pedestrian motion
  prediction in urban intersections,'' \emph{CoRR}, vol. abs/1806.09453, 2018.

\bibitem{Ziebart2008}
\BIBentryALTinterwordspacing
B.~D. Ziebart, A.~Maas, J.~A. Bagnell, and A.~K. Dey, ``Maximum entropy inverse
  reinforcement learning,'' in \emph{Proceedings of the 23rd National
  Conference on Artificial Intelligence - Volume 3}, ser. AAAI'08.\hskip 1em
  plus 0.5em minus 0.4em\relax AAAI Press, 2008, pp. 1433--1438. [Online].
  Available: \url{http://dl.acm.org/citation.cfm?id=1620270.1620297}
\BIBentrySTDinterwordspacing

\bibitem{Ziebart2009}
B.~D. Ziebart, N.~Ratliff, G.~Gallagher, C.~Mertz, K.~Peterson, J.~A. Bagnell,
  M.~Hebert, A.~K. Dey, and S.~Srinivasa, ``Planning-based prediction for
  pedestrians,'' in \emph{2009 IEEE/RSJ International Conference on Intelligent
  Robots and Systems}, Oct 2009, pp. 3931--3936.

\bibitem{Varshneya2017}
D.~Varshneya and G.~Srinivasaraghavan, ``Human trajectory prediction using
  spatially aware deep attention models,'' \emph{CoRR}, vol. abs/1705.09436,
  2017.

\bibitem{Hochreiter1997}
\BIBentryALTinterwordspacing
S.~Hochreiter and J.~Schmidhuber, ``Long short-term memory,'' \emph{Neural
  Computation}, vol.~9, no.~8, pp. 1735--1780, 1997. [Online]. Available:
  \url{https://doi.org/10.1162/neco.1997.9.8.1735}
\BIBentrySTDinterwordspacing

\bibitem{Sun2018}
L.~Sun, Z.~Yan, S.~M. Mellado, M.~Hanheide, and T.~Duckett, ``3dof pedestrian
  trajectory prediction learned from long-term autonomous mobile robot
  deployment data,'' \emph{2018 IEEE International Conference on Robotics and
  Automation (ICRA)}, pp. 1--7, 2018.

\bibitem{Alahi2016}
\BIBentryALTinterwordspacing
A.~Alahi, K.~Goel, \emph{et~al.}, \emph{Social LSTM: Human Trajectory
  Prediction in Crowded Spaces}.\hskip 1em plus 0.5em minus 0.4em\relax IEEE,
  Jun 2016. [Online]. Available: \url{http://dx.doi.org/10.1109/CVPR.2016.110}
\BIBentrySTDinterwordspacing

\bibitem{Vemula2018}
A.~Vemula, K.~Muelling, and J.~Oh, ``Social attention: Modeling attention in
  human crowds,'' \emph{2018 IEEE International Conference on Robotics and
  Automation (ICRA)}, pp. 1--7, 2018.

\bibitem{Gupta_2018_CVPR}
A.~Gupta, J.~Johnson, L.~Fei-Fei, S.~Savarese, and A.~Alahi, ``Social gan:
  Socially acceptable trajectories with generative adversarial networks,'' in
  \emph{The IEEE Conference on Computer Vision and Pattern Recognition (CVPR)},
  2018.

\bibitem{Pfeiffer2018}
M.~Pfeiffer, G.~Paolo, H.~Sommer, J.~Nieto, R.~Siegwart, and C.~Cadena, ``A
  data-driven model for interaction-aware pedestrian motion prediction in
  object cluttered environments,'' in \emph{2018 IEEE International Conference
  on Robotics and Automation (ICRA)}, May 2018, pp. 1--8.

\bibitem{Sadeghian2018}
A.~{Sadeghian}, V.~{Kosaraju}, A.~{Sadeghian}, N.~{Hirose}, S.~H.
  {Rezatofighi}, and S.~{Savarese}, ``{SoPhie: An Attentive GAN for Predicting
  Paths Compliant to Social and Physical Constraints},'' \emph{ArXiv e-prints},
  June 2018.

\bibitem{Manh2018}
H.~{Manh} and G.~{Alaghband}, ``{Scene-LSTM: A Model for Human Trajectory
  Prediction},'' \emph{ArXiv e-prints}, Aug. 2018.

\bibitem{Nikhil2018}
N.~{Nikhil} and B.~{Tran Morris}, ``{Convolutional Neural Network for
  Trajectory Prediction},'' \emph{ArXiv e-prints}, Sept. 2018.

\bibitem{radwan2018multimodal}
N.~Radwan, A.~Valada, and W.~Burgard, ``Multimodal interaction-aware motion
  prediction for autonomous street crossing,'' \emph{arXiv preprint
  arXiv:1808.06887}, 2018.

\bibitem{chandra2019traphic}
R.~Chandra, U.~Bhattacharya, A.~Bera, and D.~Manocha, ``Traphic: Trajectory
  prediction in dense and heterogeneous traffic using weighted interactions,''
  in \emph{Proceedings of the IEEE Conference on Computer Vision and Pattern
  Recognition}, 2019, pp. 8483--8492.

\bibitem{luo2018fast}
W.~Luo, B.~Yang, and R.~Urtasun, ``Fast and furious: Real time end-to-end 3d
  detection, tracking and motion forecasting with a single convolutional net,''
  in \emph{IEEE Conference on Computer Vision and Pattern Recognition (CVPR)},
  2018, pp. 3569--3577.

\bibitem{casas2018intentnet}
S.~Casas, W.~Luo, and R.~Urtasun, ``Intentnet: Learning to predict intention
  from raw sensor data,'' in \emph{Conference on Robot Learning (CoRL)}, 2018.

\bibitem{alexnet}
A.~Krizhevsky, I.~Sutskever, and G.~E. Hinton, ``{ImageNet} classification with
  deep convolutional neural networks,'' in \emph{Advances in neural information
  processing systems}, 2012, pp. 1097--1105.

\bibitem{vgg}
K.~Simonyan and A.~Zisserman, ``Very deep convolutional networks for
  large-scale image recognition,'' \emph{arXiv preprint arXiv:1409.1556}, 2014.

\bibitem{resnet}
K.~He, X.~Zhang, S.~Ren, and J.~Sun, ``Deep residual learning for image
  recognition,'' in \emph{Proceedings of the IEEE conference on computer vision
  and pattern recognition}, 2016, pp. 770--778.

\bibitem{MobileNet}
\BIBentryALTinterwordspacing
A.~G. Howard, M.~Zhu, B.~Chen, D.~Kalenichenko, W.~Wang, T.~Weyand,
  M.~Andreetto, and H.~Adam, ``Mobilenets: Efficient convolutional neural
  networks for mobile vision applications,'' \emph{CoRR}, vol. abs/1704.04861,
  2017. [Online]. Available: \url{http://arxiv.org/abs/1704.04861}
\BIBentrySTDinterwordspacing

\bibitem{ShuffleNet}
X.~Zhang, X.~Zhou, M.~Lin, and J.~Sun, ``Shufflenet: An extremely efficient
  convolutional neural network for mobile devices,'' in \emph{The IEEE
  Conference on Computer Vision and Pattern Recognition (CVPR)}, June 2018.

\bibitem{MobileNetV2}
M.~Sandler, A.~Howard, M.~Zhu, A.~Zhmoginov, and L.-C. Chen, ``Mobilenetv2:
  Inverted residuals and linear bottlenecks,'' in \emph{The IEEE Conference on
  Computer Vision and Pattern Recognition (CVPR)}, June 2018.

\bibitem{MnasNet}
\BIBentryALTinterwordspacing
M.~Tan, B.~Chen, R.~Pang, V.~Vasudevan, and Q.~V. Le, ``Mnasnet: Platform-aware
  neural architecture search for mobile,'' \emph{CoRR}, vol. abs/1807.11626,
  2018. [Online]. Available: \url{http://arxiv.org/abs/1807.11626}
\BIBentrySTDinterwordspacing

\bibitem{ZophL16}
\BIBentryALTinterwordspacing
B.~Zoph and Q.~V. Le, ``Neural architecture search with reinforcement
  learning,'' \emph{CoRR}, vol. abs/1611.01578, 2016. [Online]. Available:
  \url{http://arxiv.org/abs/1611.01578}
\BIBentrySTDinterwordspacing

\bibitem{ShuffleNetV2}
\BIBentryALTinterwordspacing
N.~Ma, X.~Zhang, H.~Zheng, and J.~Sun, ``Shufflenet {V2:} practical guidelines
  for efficient {CNN} architecture design,'' \emph{CoRR}, vol. abs/1807.11164,
  2018. [Online]. Available: \url{http://arxiv.org/abs/1807.11164}
\BIBentrySTDinterwordspacing

\bibitem{wan2000unscented}
E.~A. Wan and R.~Van Der~Merwe, ``The unscented kalman filter for nonlinear
  estimation,'' in \emph{Adaptive Systems for Signal Processing,
  Communications, and Control Symposium 2000. AS-SPCC. The IEEE 2000}.\hskip
  1em plus 0.5em minus 0.4em\relax Ieee, 2000, pp. 153--158.

\bibitem{tensorflow2015-whitepaper}
\BIBentryALTinterwordspacing
M.~Abadi, A.~Agarwal, P.~Barham, E.~Brevdo, \emph{et~al.}, ``{TensorFlow}:
  Large-scale machine learning on heterogeneous systems,'' 2015. [Online].
  Available: \url{https://www.tensorflow.org/}
\BIBentrySTDinterwordspacing

\bibitem{sergeev2018horovod}
A.~Sergeev and M.~D. Balso, ``Horovod: fast and easy distributed deep learning
  in tensorflow,'' \emph{arXiv preprint arXiv:1802.05799}, 2018.

\bibitem{kingma2014adam}
D.~P. Kingma and J.~Ba, ``Adam: A method for stochastic optimization,''
  \emph{arXiv preprint arXiv:1412.6980}, 2014.

\end{thebibliography}
}

\end{document}